\documentclass{article} 
\usepackage{iclr2027_conference,times}

\usepackage{amsmath,amsfonts,bm}

\def\eqref#1{equation~\ref{#1}}

\def\1{\bm{1}}

\DeclareMathAlphabet{\mathsfit}{\encodingdefault}{\sfdefault}{m}{sl}
\SetMathAlphabet{\mathsfit}{bold}{\encodingdefault}{\sfdefault}{bx}{n}

\usepackage{hyperref}
\usepackage{url}
\usepackage{algorithm}
\usepackage{algorithmic}
\usepackage{comment}
\usepackage{amsmath}
\usepackage{amssymb}
\usepackage{pifont}
\usepackage{subcaption}
\usepackage{cleveref}
\usepackage{multirow}
\usepackage{graphicx} 
\usepackage{booktabs}
\usepackage{wrapfig}
\usepackage{newfloat}
\usepackage{listings}
\usepackage{tabularx}
\usepackage{array}
\usepackage{ragged2e}
\renewcommand{\headrulewidth}{0pt}

\DeclareCaptionStyle{ruled}{labelfont=normalfont,labelsep=colon,strut=off} 
\floatstyle{ruled}
\newfloat{listing}{tb}{lst}{}
\floatname{listing}{Listing}

\title{RA-CAD: Learning Post-Execution Critique for State-Aware Text-to-CAD Generation}

\author{Shuhao Yan\textsuperscript{\rm 1}\quad
    Changhao He\textsuperscript{\rm 1}\quad
    Peng Hu\textsuperscript{\rm 1,\rm 3}\thanks{Corresponding author.}\quad
    Xi Peng\textsuperscript{\rm 2,\rm 3}\\
    College of Computer Science, Sichuan University, Chengdu, China.\textsuperscript{\rm 1}\\
    School of Artificial Intelligence, Sichuan University, Chengdu, China.\textsuperscript{\rm 2}\\
    National Key Laboratory of Fundamental Algorithms and Models for Engineering \\Numerical Simulation, Sichuan University, Chengdu, China.\textsuperscript{\rm 3}
}

\iclrfinalcopy 

\begin{document}

\maketitle

\begin{abstract}
Text-to-CAD generation translates natural-language design intent into editable and executable parametric computer-aided design (CAD) codes, reducing the expertise and effort required for manual modeling. Existing methods incorporate fixed, externally supplied, prompt-induced, or separately optimized critique mechanisms to optimize the generation process, but they do not necessarily optimize how feedback is interpreted and translated into effective corrective actions throughout the generation process. To bridge this feedback-utilization gap, we present RA-CAD (ReAct Agent for CAD), a state-aware agent that interacts with the CAD environment through a Generate--Execute--Critique--Rewrite loop. At each iteration, RA-CAD executes the current code and observes its outcome. Conditioned on the design instruction, current code, and execution feedback, the agent then generates an explicit post-execution critique as an intermediate policy action. This critique either validates the current result for termination or provides revision-oriented guidance that conditions the next rewrite. CAD Code Bootstrapping (CCB) first establishes fundamental parametric CAD coding capabilities through supervised fine-tuning. Feedback-Driven Agent Optimization (FAO) subsequently applies trajectory-level Group Relative Policy Optimization to both policy-generated code and critique sequences, assigning terminal F1 and Chamfer Distance rewards to the complete interaction trajectory. This formulation makes critique an outcome-aligned, learnable policy decision rather than an unoptimized auxiliary output. Experiments on CADFusion and Text2CAD show that RA-CAD achieves state-of-the-art execution validity and geometric quality compared with existing methods and strong proprietary language models, demonstrating the effectiveness of the proposed state-aware text-to-CAD agent.
\end{abstract}


\section{Introduction}

\begin{wrapfigure}[19]{r}{0.5\columnwidth}
\centering
\includegraphics[width=0.48\columnwidth]{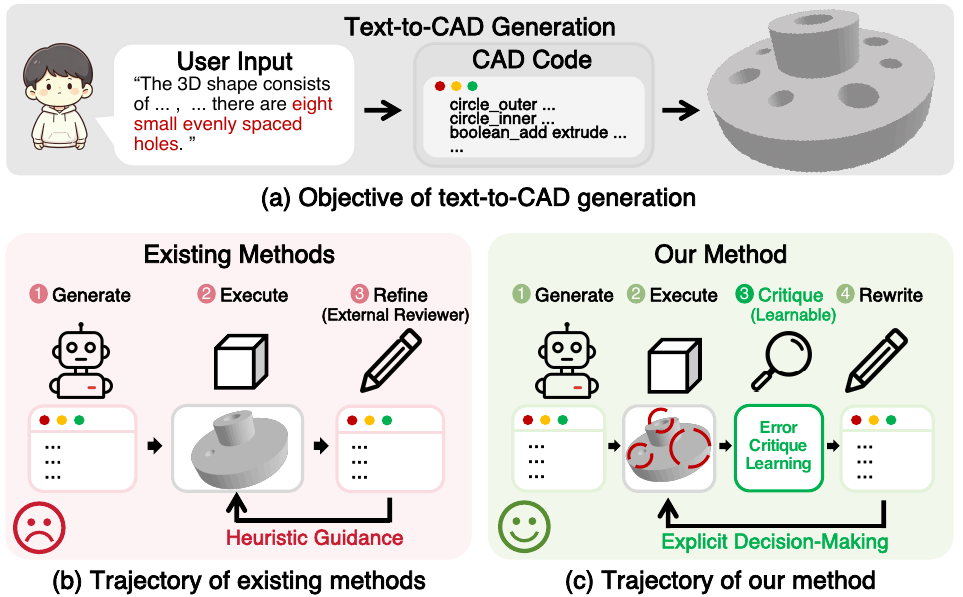}
\caption{Comparison of CAD generation methods between existing methods and our method. Existing methods treat feedback as heuristic guidance from external visual reviewers. Our work models it as an explicit decision-making process with an agent-side component.}
\label{Motivation}
\end{wrapfigure}

Computer-aided design (CAD) represents engineered objects through precise, editable geometric operations and is central to product design and manufacturing~\citep{wu2021deepcad, xu2022skexgen}. Creating these models manually, however, requires substantial domain expertise and effort~\citep{deng2023whatsets}. In recent years, various automatic CAD generation technologies have been proposed to translate user specifications or design intent into executable, editable CAD representations as shown in \Cref{Motivation}.(a).

Early neural approaches learned unconditional distributions over CAD construction sequences~\citep{wu2021deepcad,xu2023hnc}. Conditional methods subsequently emerged to guide generation with natural-language descriptions, typically formulating the task as one-shot supervised prediction~\citep{khan2024text2cad,he2025cadcoder,xie2025texttocadquery,yuan2026procad}. Preference learning and geometric-reward optimization further improve final-output fidelity~\citep{wang2025texttocad,guan2025cadcoder,li2026recad}. Nevertheless, a complete CAD program is brittle: a single invalid parameter or inconsistent operation may prevent execution, while syntactically valid code may still instantiate the wrong geometry. A one-shot decoder cannot use these outcomes to repair its prediction.

Recent systems address this limitation through execution feedback, external visual reviewers, tool interaction, or iterative refinement \citep{li2026seekcad,gong2026toolcad,hu2026itercad,fan2026caddesigner,liu2026cadfactory}. The availability of feedback, however, does not by itself specify how feedback utilization is represented or learned: it may serve only as a training reward, be produced by a fixed reviewer or compiler, or remain embedded in a broader reasoning trace. Regardless of the form, they treat it as heuristic guidance as shown in \Cref{Motivation}.(b), and its practical value hinges on a single critical post-execution decision: judging whether the instantiated code satisfies the instruction, identifying actionable defects, and deciding whether and how to revise it. In CAD, an inaccurate decision can cause premature acceptance, unnecessary edits, or accumulated syntactic and geometric errors. Reliable generation therefore requires executable code synthesis and a post-execution critique decision that is optimized together with rewriting.

\begin{figure*}[t]
\centering
\includegraphics[width=\textwidth]{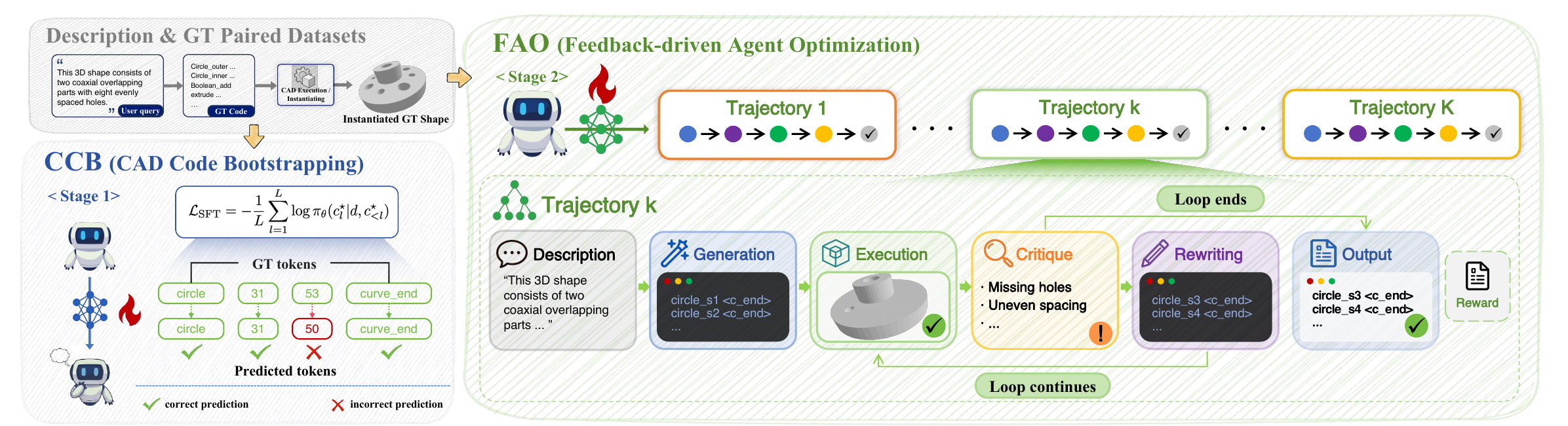} 
\caption{Overview of our training framework, including two stages: CAD Code Bootstrapping (CCB) and Feedback-driven Agent Optimization (FAO). In the CCB stage, the agent is initialized by supervised fine-tuning using paired text-CAD data to learn the syntax and geometric constraints of parametric CAD codes. In the FAO stage, the GRPO algorithm is adopted to fine-tune the complete execution trajectory of the agent, enabling the generation and critique policy to co-evolve under feedback-driven execution.}
\label{Pipeline}
\end{figure*}

RA-CAD addresses this problem with a ReAct Agent \citep{yao2022react} that operates as a state-aware Generate--Execute--Critique--Rewrite loop. As shown in \Cref{Pipeline}, RA-CAD organizes CAD generation as a critique-guided revision process under environment interaction. Rather than treating critique as a fixed post-hoc prompt or an auxiliary textual analysis step, we model it as an agent-side decision component that produces diagnostic signals for subsequent rewriting and termination. Specifically, CAD Code Bootstrapping (CCB) initializes code generation with supervised fine-tuning (SFT), whereas Feedback-Driven Agent Optimization (FAO) uses terminal code and geometry rewards to optimize all policy-generated code and critique tokens through Group Relative Policy Optimization (GRPO). RA-CAD then explicitly factorizes post-execution critique as a learnable policy component and jointly optimizes critique and code proposal/rewrite over complete environment-interaction trajectories as shown in \Cref{Motivation}.(c), rather than an external reviewer as shown in \Cref{Motivation}.(b). In this way, RA-CAD learns not only to propose CAD code, but also to critique intermediate results and use those critique outputs to guide later revisions. By optimizing the full interaction trajectory, the framework improves the executability, geometric accuracy, and robustness of complex CAD code generation.

The main contributions are summarized as follows:
\begin{itemize}
\item We formulate state-aware text-to-CAD generation as an agent interaction process with an explicitly factorized, learnable post-execution critique policy. The critique is explicitly modeled as an agent-side decision component rather than treated as a fixed prompt template or a purely post-hoc analysis step.
\item We propose trajectory-level FAO with CCB, which assigns terminal CAD-quality rewards to complete interaction trajectories and jointly optimizes code generation, post-execution critique, and rewriting.
\item Experimental results on CADFusion and Text2CAD demonstrate that closed-loop, critique-guided revision substantially improves execution validity and geometric quality over existing text-to-CAD methods and strong proprietary language models.
\end{itemize}

\section{Related Work}
\subsection{CAD Generation}
Sequence models such as DeepCAD~\citep{wu2021deepcad}, SkexGen~\citep{xu2022skexgen}, HNC-CAD~\citep{xu2023hnc}, SketchGen~\citep{para2021sketchgen}, and CAD-as-Language~\citep{ganin2021cadaslanguage} learn structured sketch or sketch-extrude representations, whereas BRepGen~\citep{xu2024brepgen} and SolidGen~\citep{jayaraman2023solidgen} directly model boundary representations. Methods conditioned on images/sketches~\citep{li2022free2cad, you2025img2cad, chen2025cadcrafter, doris2026cadcoder}, or point clouds~\citep{khan2024cadsignet, liu2024point2cad, rukhovich2025cadrecoder} reconstruct CAD codes from geometric observations; FlexCAD and GeoCAD additionally support controllable editing~\citep{zhang2025flexcad,zhang2026geocad}. CADReview defines CAD review as an independent task \citep{chen2025cadreview}, but it is aimed at error detection and code correction for existing CAD code, lacking the ability to generate CAD models from natural language. Although these inputs provide strong geometric constraints, natural language offers a more direct interface for expressing high-level design intent. 

Text-to-CAD methods instead predict parametric sequences or CAD code from descriptions~\citep{yavartanoo2024text2cad,xie2025texttocadquery, Li2025cadllama,govindarajan2025cadmium}. Most emphasize supervised mapping, planning, or controllable generation~\citep{yang2026intent2cad,liu2026expconcad,pang2026goprcad,li2026cadwam}. Their outputs may be executed for evaluation, but execution is not necessarily represented as a learned post-execution critique action within a refinement trajectory.

\subsection{Feedback-Guided CAD Refinement}
Feedback-guided methods differ in both the source and representation of feedback. CADFusion~\citep{wang2025texttocad} pioneered the introduction of DPO~\citep{rafailov2023dpo}, using rendered 3D visual feedback as preference signals to improve geometric consistency and visual fidelity.
In order to further improve geometric accuracy, CAD-Coder~\citep{guan2025cadcoder} introduced GRPO~\citep{shao2024deepseekmath}, which further advanced geometric accuracy with GRPO, leveraging chain-of-thought (CoT)~\citep{wei2022cot} inference and geometric metrics as optimization signals.
ReCAD~\citep{li2026recad}, PR-CAD~\citep{an2026prcad}, CME-CAD~\citep{niu2026cmecad}, ToolCAD~\citep{gong2026toolcad}, IterCAD~\citep{hu2026itercad}, PhyCAD~\citep{cui2026phycad}, CRAFT~\citep{rafi2026craft}, and CIT-CAD~\citep{du2026citcad} further demonstrate the value of reinforcement learning, structured reasoning, multi-turn refinement, and environment interaction in CAD generation. 
Although these methods show that feedback can improve CAD generation, they primarily optimize structured inputs, generation policies, or tool-use processes. In contrast, we formulate text-to-CAD generation as a closed-loop refinement process driven by explicit learnable post-execution critique and jointly optimize the generation, rewriting, and critique policies at the trajectory level.

\section{Method}
\subsection{Problem Formulation}
\begin{figure*}[t]
\centering
\includegraphics[width=\textwidth]{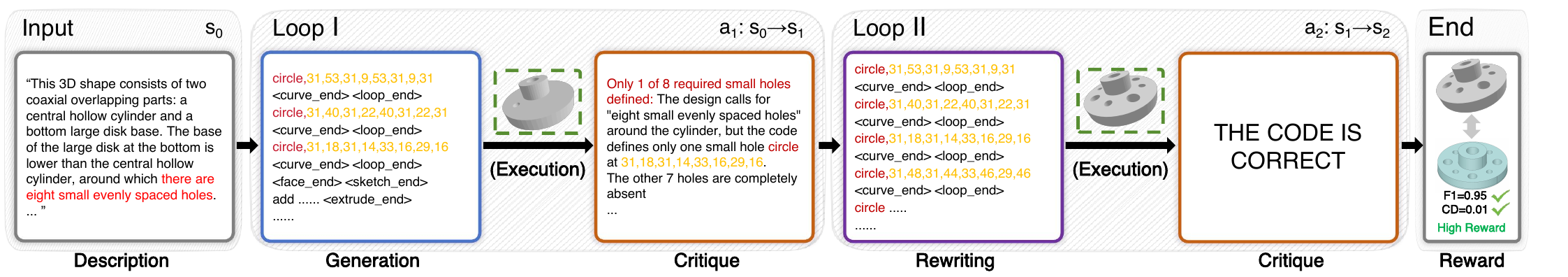} 
\caption{Overview of the agent trajectory in RA-CAD. The input description is considered the initial state $s_0$. At each loop, the agent executes action $a_t$ and interacts with the CAD environment, producing a state transition from $s_{t-1}$ to $s_t$.}
\label{Traj}
\end{figure*}

Let $\mathcal D=\{(d^{(i)},c^{\star(i)})\}_{i=1}^{N}$ be a text-to-CAD dataset, where $d$ is a natural-language design description and $c^\star=(c^\star_1,\ldots,c^\star_L)$ is its ground-truth parametric CAD code. A CAD executor $\mathcal E$ maps a candidate code $c$ to an execution result $e=\mathcal E(c)$, which records whether the code is valid and any available diagnostic message. If execution succeeds, the CAD environment instantiates geometry $\mathcal G(c)$. Text-to-CAD generation therefore seeks a conditional policy
\begin{equation}
  \hat c\sim\pi_\theta(\,\cdot\mid d),\qquad \hat e=\mathcal E(\hat c),
\end{equation}
where $\hat c$ is executable, agrees with the design intent in $d$, and instantiates geometry consistent with $\mathcal G(c^\star)$. 
A one-shot policy ends after producing $\hat c$ and cannot use $\hat e$ to repair the code. RA-CAD instead formulates generation as a finite sequence of interactions between the policy and $\mathcal E$. Starting from a state initialized by $d$, round $t$ follows the high-level transition
\begin{equation}
  \begin{aligned}
    \tilde c_t&\sim\pi_\theta^{\mathrm{g\&r}}(\,\cdot\mid s_{t-1}),
    &e_t&=\mathcal E(\tilde c_t),\\
    \tilde f_t&\sim\pi_\theta^{\mathrm{crit}}(\,\cdot\mid\bar s_t),
  \end{aligned}
\end{equation}
where $\tilde c_t$ is the initial code at $t=1$ or a rewrite at $t>1$, $\bar s_t$ is the state after execution, and $\tilde f_t$ is the explicit post-execution critique. The critique either accepts the current code or supplies guidance that conditions the next rewrite. The process terminates upon acceptance or at the interaction limit and returns the final code $c_T$. As shown in \Cref{Traj}, its complete trajectory is
\begin{equation}
  \tau=(s_0,a_1,s_1,\ldots,a_T,s_T),
\end{equation}
where $\tau$ represents the trajectory, $s_0$ represents the initial state consisting solely of $d$, $a_{>0}$ indicates each composite action, $s_{>0}$ represents the each complete state, and $T$ represents the final round.

As shown in \Cref{Pipeline}, RA-CAD trains this policy in two stages. CAD Code Bootstrapping (CCB) first learns the description-to-code mapping from $(d,c^\star)$ pairs by supervised fine-tuning. Feedback-driven Agent Optimization (FAO) then samples complete Generate--Execute--Critique--Rewrite trajectories and applies terminal CAD-quality rewards to all policy-generated code and critique sequences. The following subsections specify these two stages and the trajectory-level objective.

\subsection{CAD Code Bootstrapping}
Strict CAD syntax and geometric constraints make it unreliable for reinforcement learning to learn valid CAD generation from scratch. CCB therefore first performs supervised fine-tuning on the basic language model to provide a lightweight cold-start that exposes the model to basic CAD syntax and parametric code structure.

During the training process, we adopt a teacher forcing policy to gradually predict each token in the target CAD code through autoregression (next-token prediction). In the $l$-th generation step, the model predicts the current token $c^\star_l$ based on the previous $l-1$ ground-truth tokens as conditions, and uses cross-entropy loss for supervised optimization. The entire training objective is defined as
\begin{equation}
\mathcal{L}_{\mathrm{SFT}} = -\frac{1}{L} \sum_{l=1}^{L} \log \pi_{\theta}(c^\star_l|d,c^\star_{<l}),
\end{equation}
where $L$ represents the length of the ground-truth code, $\pi_\theta$ denotes the model policy with parameters $\theta$.

\subsection{Feedback-driven Agent Optimization}
While CAD Code Bootstrapping (CCB) enables the agent to generate executable CAD code, it does not explicitly optimize the agent's reasoning and decision-making ability during multi-round interaction. To address this limitation, we design a ReAct Agent that works as a closed-loop CAD generation process, and we model the process as a Markov Decision Process (MDP) \citep{bellman1957mdp, li2019overviewmdp}, in which the agent makes decisions conditioned on the current design objective, historical generation result, and execution result from the environment. The complete process forms a trajectory $\tau$, which is optimized according to the quality of the final executable CAD model, as shown in \Cref{Traj}. In this way, GRPO optimization is no longer focused on single CAD code prediction, but on the decision-making policy of the entire agent.

\subsubsection{State}
To capture both the current CAD generation result and the execution result returned by the environment, we define the state at round $t$ as
\begin{equation}
s_t = (d, c_t, e_t, f_t),
\end{equation}
where $d$ denotes the user-provided text description, $c_t$ denotes the current CAD code candidate, $e_t$ denotes the results returned by the CAD execution environment, and $f_t$ denotes the critique result associated with the current round. Therefore, each decision made by the agent is conditioned on the design objective, the current code state, the execution result, and the critique feedback.

\subsubsection{Action}
Our framework contains four functional modules: Generation, Execution, Critique, and Rewriting. Among them, Generation, Critique, and Rewriting constitute the agent-side decision process, while Execution belongs to the environment and is responsible for executing the current CAD code and returning feedback.

Instead of treating Generation, Critique, and Rewriting as isolated actions, we formulate the agent's decision at each round as a composite action:
\begin{equation}
a_t = (\tilde{c}_t, \tilde{f}_t),
\end{equation}
where $a_t$ is temporally ordered composite decision, $\tilde{c}_t$ denotes the CAD code proposed by the agent at the current round which depends on the action mode of $m_t \in \{\texttt{generate}, \texttt{rewrite}\}$, and $\tilde{f}_t$ denotes the critique output, which is the evaluation of whether the current code and its execution result meet the design objectives.
Specifically, when $m_1=\texttt{generate}$, the agent only generates the initial CAD code based on the input description. When $m_{>1}=\texttt{rewrite}$, the agent will modify the previous CAD code based on the text description, execution results, and previous evaluation records. After getting the generated code in the above steps and then obtaining the execution result, the agent critiques it again. On this basis, we can define the policy in detail as
\begin{equation}
\pi_\theta(a_t\mid s_{t-1})
=
\pi_\theta^{\mathrm{g\&r}}(\tilde c_t\mid s_{t-1})
\pi_\theta^{\mathrm{crit}}
(\tilde f_t\mid \Bar{s}_t).
\end{equation}
where $\pi_\theta^{\mathrm{g\&r}}$ and $\pi_\theta^{\mathrm{crit}}$ denote the conditional distributions for Generation/Rewriting and Critique, respectively. After $\tilde{c}_t$ is produced, the CAD environment updates to intermediate state $\bar{s}_t$, which has been updated to $\tilde{c}_t$ and $e_t$, respectively. The Critique module then produces $\tilde{f}_t$ conditioned on $\bar{s}_t$. As shown in \Cref{Traj}, after fully executing action $a_t$, the environment state transitions from $s_{t-1} $ to $s_t$.

The four modules are described as follows:

\textbf{1. Generation}
The Generation module is responsible for producing the initial CAD code from the input text description. Given an initial state $s_0$ that only contains $d$, the Generation module will generate the initial candidate CAD code $c_1$ and pass it into the environment. Rather than being directly treated as the final output, the generated code is regarded as an intermediate hypothesis for the current design objective, which will be further verified and refined through interaction with the execution environment.

\textbf{2. Execution}
The Execution module belongs to the environment rather than the agent itself. It instantiates and executes the CAD code $c_t$ proposed by the agent, detects syntax errors, invalid parameters, and other execution failures, and returns execution status together with the corresponding error message, if it exists, to the environment. Then it forms an intermediate state $\bar s_{t}$.

\textbf{3. Critique}
The Critique module evaluates the current CAD result by jointly considering the intermediate state $\bar s_{t}$ (including input description $d$, the generated CAD code $c_t$, and the execution result $e_t$). It determines whether the current code satisfies the design requirements in terms of intent consistency, geometric validity, and overall model quality. The critique output not only decides whether further optimization is needed, but also provides revision suggestions for the next round.

\textbf{4. Rewriting}
When the Critique results indicate that the current code still has room for improvement, the rewriting module will work as the start of a new round of actions for the agent. It will rewrite the code based on the complete state $s_{t} = (d, c_{t}, e_{t}, f_{t})$ updated from the previous round of actions. The modified code $c_{t+1}$ will then be sent back to the Execution module for verification. The interaction loop continues until the critique module determines that the design goal has been satisfied or the maximum number of iterations is reached.

\subsubsection{Reward}
Since the ultimate goal of CAD codes is to generate parameterized models that meet design requirements and can be executed correctly, we calculate the final reward value for the complete trajectory based on the final generated code results.
For the final generated result, we comprehensively consider the following aspects: how close the generated sequence is to the ground-truth sequence; whether the CAD code can be successfully executed; and geometric similarity between the generated model and ground-truth model. Therefore, the reward function can be expressed as
\begin{equation}
R = \lambda_1R_{\text{F1}} + \lambda_2R_{\text{CD}},
\end{equation}
where $\lambda_1, \lambda_2 $ are weight coefficients; $R_{\text{F1}}$ is the F1 score reward of the basic sketch topological tokens, which measures the similarity between the generated parameter sequence and the real parameter sequence; $R_{\text{CD}}$ is the distance of the instantiated model of the final result of the trajectory, which not only measures whether the code is successfully executed, but also measures the quality of the executable model generation. Specifically, $R_{\text{CD}}$ and $R_{\text{F1}}$ are defined as: 
\begin{equation}
R_{\text{CD}} = 
\begin{cases}
e^{-\gamma \text{CD}}, & \text{if CAD code is executable} \\
0, & \text{otherwise}
\end{cases},
\end{equation}
\begin{equation}
R_{\text{F1}} = R_{\text{Avg F1}} = (\text{F1}_{\text{line}} + \text{F1}_{\text{arc}} + \text{F1}_{\text{circle}})/3,
\end{equation}
where $\gamma $ is hyperparameter and $\text{CD}$ is the chamfer distance between the point cloud corresponding to the generated CAD model $\mathcal G(c)$ and the point cloud corresponding to the ground-truth CAD model $\mathcal G(c^\star)$, and $R_{\text{F1}}$ only considers the basic sketch  topological tokens $\{\texttt{line}, \texttt{arc}, \texttt{circle}\}$ as they dominate CAD sequences and determine structural correctness. Extrusion tokens $\{\texttt{add}, \texttt{cut}, \texttt{intersect}\}$ exhibit substantially higher numerical variability than sketch topology tokens, making token-level matching less reliable, with their quality indirectly measured by the CD reward. Due to the dependence of rewards on the entire agent inference process, they cannot be decomposed into a single step, but rather serve as the final reward for the entire trajectory. Although the reward is defined on the final trajectory outcome, the critique behavior is implicitly optimized because critique outputs constitute part of the action trajectory updated by GRPO.

\subsubsection{Policy Optimization with GRPO}
Given the trajectory-level reward \( R \), we adopt GRPO to optimize the agent policy \( \pi_\theta \). It normalizes rewards within each group of trajectories sampled from the same input, providing a stable and computation-efficient advantage estimation. Firstly, it calculates the group-relative advantage $A_k[p]$ of the $k$-th trajectory of the $i$-th sample through the reward value, which is defined as follows
\begin{equation} 
A_k[p] = \displaystyle \frac{R_k - \mu_i}{\sigma_i},
\end{equation}
where $R_k$ is the final reward of the $k$-th trajectory, $\mu_i$ represents group-wise mean, $\sigma_i$ represents standard deviation, and $p$ is the valid token position in the trajectory. Given the advantages, we compute the token-level probability ratio 
\begin{equation} 
\rho_k[p] = \frac{\pi_\theta(l_{k,t}[p] \mid s_{k,t},\; l_{k,t}[:p])}{\pi_{\theta_{\text{old}}}(l_{k,t}[p] \mid s_{k,t},\; l_{k,t}[:p])} 
\end{equation}
between the current policy \(\pi_\theta\) and the old policy \(\pi_{\theta_{\text{old}}}\), where $l_{k,t} \in \{l_{k,1},\dots,l_{k, T} \}$ represents the output sequence in the $t$-th sub-action round, including the generated code sequence and the critique sequence. Then we optimize the clipped surrogate objective defined as 
\begin{equation}
\mathcal{L}_{\text{GRPO}} = 
\begin{aligned}[t]
&-\frac{1}{K} \sum_{k=1}^{K} 
\begin{aligned}[t] 
&\frac{1}{L_k} \sum_{p=0}^{L_k-1} \min\big( \rho_k[p] \cdot A_k[p],\; \operatorname{clip}(\rho_k[p],\; 1-\varepsilon_{-},\; 1+\varepsilon_{+}) \cdot A_k[p] )
\end{aligned}\\
&+ \beta \cdot \mathbb{E}_{k,p}\big( \log\frac{\pi_\theta}{\pi_{\text{ref}}} - 1 + \frac{\pi_{\text{ref}}}{\pi_\theta}) ,
\end{aligned}
\end{equation}
where \(K\) is the total number of trajectories, \(L_k\) is the length of the \(k\)-th trajectory, \(\varepsilon_{-}\) and \(\varepsilon_{+}\) are the clipping hyperparameter that constrains the magnitude of policy updates to ensure training stability, and $\beta \cdot \mathbb{E}_{k,p}(\cdot)$ are KL regularization and its corresponding hyperparameter. By optimizing this objective with GRPO, the generation and critique policy within the agent are jointly and synergistically optimized under the guidance of the state.

\section{Experiments}
In this section, we conducted experiments on the CADFusion and Text2CAD datasets to comprehensively evaluate the effectiveness of the proposed method. Firstly, the experimental setups are introduced, including the datasets, training settings, and evaluation metrics. Subsequently, RA-CAD will be compared with existing representative methods and mainstream proprietary LLMs to verify its overall performance. After that, we analyze the contribution of each core module to the model performance through ablation experiments, further verifying the effectiveness of the proposed framework.

\subsection{Setups}
\subsubsection{Datasets}
We adopt CADFusion's parameter sequence construction rules \citep{wang2025texttocad} to reparameterize the CAD models in the Text2CAD and CADFusion datasets and construct a unified CAD code sequence. The text description retains the natural language annotations provided by the original datasets, while performing data cleaning on the datasets to remove samples whose ground-truth parameter sequences cannot be instantiated, in order to ensure data quality and training stability. 
It should be noted that when processing the Text2CAD dataset, considering that user descriptions in actual CAD design scenarios typically focus on high-level semantics rather than the specific geometric parameters, we only used its intermediate-level text description for experimentation.
The processed CADFusion and Text2CAD datasets contain approximately 20k and 60k samples, respectively, and the ratio of training, validation, and testing sets is retained at 90\%-5\%-5\%.

\subsubsection{Implementation Details}
We use the open-source Meta-Llama-3-8B-Instruct as the base model. In the CCB stage, the CADFusion SFT training framework is used, and Low Rank Adaptation (LoRA) is used for efficient parameter tuning. The model is trained for 9 epochs, with a learning rate set to $1\times10^{-4}$, LoRA hyperparameters set to $r=32$, $\alpha=32$, and an optimizer using AdamW. In the FAO stage, we train based on the AgentLightning framework \citep{luo2025agentlightning} and implement distributed training using Fully Sharded Data Parallel (FSDP). The training learning rate is set to $1\times10^{-6}$. Each input sample has $K=8$ trajectories, and the training batch size is 16. GRPO hyperparameters set to $\lambda_1=0.1$, $\lambda_2=0.9$, $\gamma=1$, $\beta = 10^{-3}$, $\varepsilon_{-}=0.2$, $\varepsilon_{+}=0.3$. Max turn set to $T_{max}=5$. All experiments were conducted on two NVIDIA GeForce RTX 4090 (48 GB) GPUs.

\subsubsection{Baseline}
We consider three baselines, the first of which is the transformer architecture for implementing end-to-end text-to-parametric CAD model generation in Text2CAD~\citep{khan2024text2cad}. Due to significant differences between its model structure and the method proposed in this paper, we follow the original implementation. The second type is CADFusion's~\citep{wang2025texttocad} LLM fine-tuning architecture based on visual signals for preference learning. The third type is IterCAD's~\citep{hu2026itercad} multimodal agent framework for CAD generation that uses multi-view engineering drawings as spatial anchors and operates through closed-loop iteration.

\subsubsection{Metrics}
We evaluated the model from two aspects: the parameter sequence and the instantiated CAD model. The quality of the parameter sequence was measured using the evaluation protocols of Text2CAD and CADFusion, based on the F1 scores of each topological operation (including line, arc, circle, and extrusion) and the average of basic sketch topological tokens. The model quality was assessed by instantiating the generated sequence into a 3D CAD model and comparing it with the real model point cloud. The geometric accuracy, executability, and distribution consistency were evaluated using average CD (Avg CD), median CD (Med CD), and invalidity ratio (IR).

\begin{table*}[t]
\caption{Comparison of RA-CAD with existing methods on CADFusion dataset and Text2CAD dataset. Single denotes one-shot inference, Repeat-5 denotes repeat 5 turns and taking the first executable sample, and Turns-5 denotes max 5 turns. All indicators are multiplied by $10^2$ for readability. $\uparrow$: The higher the better; $\downarrow$: The lower the better. The best performance is highlighted in bold.}
\centering
\small

\begin{subtable}{\textwidth}
\setlength{\tabcolsep}{0.4mm}
\centering
\small 
\caption{CADFusion Dataset}
\begin{tabular}{l|l|l|ccccc|cc|c}
  \toprule
  Method & Setting & Venue & Line F1$\uparrow$ & Arc F1$\uparrow$ & Circle F1$\uparrow$ & Avg F1$\uparrow$ & Ext F1$\uparrow$ & Avg CD$\downarrow$ & Med CD$\downarrow$ & IR$\downarrow$ \\
  \midrule
  \multirow{2}{*}{Text2CAD} 
    & Single      & \multirow{2}{*}{NeurIPS'24} & 75.91 & 72.33 & 77.58 & 75.27 & 87.43 & 41.05 & 31.62 & 58.15 \\
    & Repeat-5    & & 77.66 & 68.58 & 79.42 & 75.22 & 87.48 & 41.75 & 33.77 & 23.66 \\
  \midrule
  \multirow{2}{*}{CADFusion} 
    & Single      & \multirow{2}{*}{ICML'25} & 81.67 & 81.42 & 87.67 & 83.59 & 90.32 & 40.76 & 29.82 & 20.82 \\
    & Repeat-5    & & 81.94 & 82.71 & 88.39 & 84.35 & 90.54 & 40.43 & 29.93 & 3.15 \\
  \midrule
  IterCAD & Turns-5 & EMNLP'26 
    & 74.35 & 79.99 & 85.30 & 79.88 & 87.71 & 38.25 & 33.02 & 1.58\\
  \midrule
  RA-CAD & Turns-5 & -- 
    & \textbf{82.51} & \textbf{82.99} & \textbf{89.62} & \textbf{85.04} & \textbf{91.43} & \textbf{29.04} & \textbf{23.77} & \textbf{1.05} \\
  \bottomrule
\end{tabular}
\end{subtable}

\begin{subtable}{\textwidth}
\setlength{\tabcolsep}{0.4mm}
\centering
\small 
\caption{Text2CAD Dataset}
\begin{tabular}{l|l|l|ccccc|cc|c}
  \toprule
  Method & Setting & Venue & Line F1$\uparrow$ & Arc F1$\uparrow$ & Circle F1$\uparrow$ & Avg F1$\uparrow$ & Ext F1$\uparrow$ & Avg CD$\downarrow$ & Med CD$\downarrow$ & IR$\downarrow$ \\
  \midrule
  \multirow{2}{*}{Text2CAD} 
    & Single      & \multirow{2}{*}{NeurIPS'24} & 76.60 & 60.94 & 76.76 & 71.43 & 84.11 & 40.99 & 33.64 & 55.93 \\
    & Repeat-5    & & \textbf{78.49} & 62.76 & 79.11 & 73.46 & 84.62 & 41.03 & 33.30 & 17.57\\
  \midrule
  \multirow{2}{*}{CADFusion} 
    & Single      & \multirow{2}{*}{ICML'25} & 74.67 & 71.97 & 81.61 & 76.08 & 84.50 & 50.94 & 40.97 & 22.43 \\
    & Repeat-5    & & 74.84 & 72.17 & 81.14 & 76.05 & 84.46 & 50.70 & 39.58 & 1.75 \\
  \midrule
  IterCAD & Turns-5 & EMNLP'26 
    & 77.99 & 72.01 & \textbf{81.48} & \textbf{77.16} & 80.95 & 37.81 & 35.11 & 1.49 \\
  \midrule
  RA-CAD & Turns-5 & -- 
    & 76.18 & \textbf{72.89} & 80.63 & 76.57 & \textbf{86.72} & \textbf{33.35} & \textbf{29.03} & \textbf{0.89} \\
  \bottomrule
\end{tabular}
\end{subtable}

\label{MainR}
\end{table*}

\subsection{Main Results}
We evaluated RA-CAD through comprehensive experiments on the CADFusion~\citep{wang2025texttocad} and Text2CAD~\citep{khan2024text2cad} datasets, and compared its final parameter sequence generation quality and instantiated CAD model quality with existing methods. The final results are shown in \Cref{MainR}. Our results indicate that our complete method achieved the best results on most metrics, outperforming previous work in terms of geometric accuracy. Specifically, for the CADFusion dataset, it reduced the Avg CD to 29.04 and IR to 1.05, and for the Text2CAD dataset, it reduced the Avg CD to 33.35 and IR to 0.89. \Cref{CaseS} exhibits the qualitative results. We can observe that, compared with the powerful baselines Text2CAD and CADFusion, RA-CAD has achieved significant improvements.
The optimization objective of RA-CAD is more inclined towards global geometric consistency rather than simply mimicking the parameter distribution of the training set. Although CADFusion is more aggressive in predicting extrusion tokens, RA-CAD achieves a better overall performance between the final rendered geometry (CD) and invalidity ratio (IR). This indicates that RA-CAD with the learnable post-execution critique mechanism understands the geometric prior that parameter combinations determine shape, rather than merely memorizing the statistical frequencies of the parameters.

\subsection{Proprietary LLMs Studies}

\begin{figure*}[t]
\centering
\includegraphics[width=\textwidth]{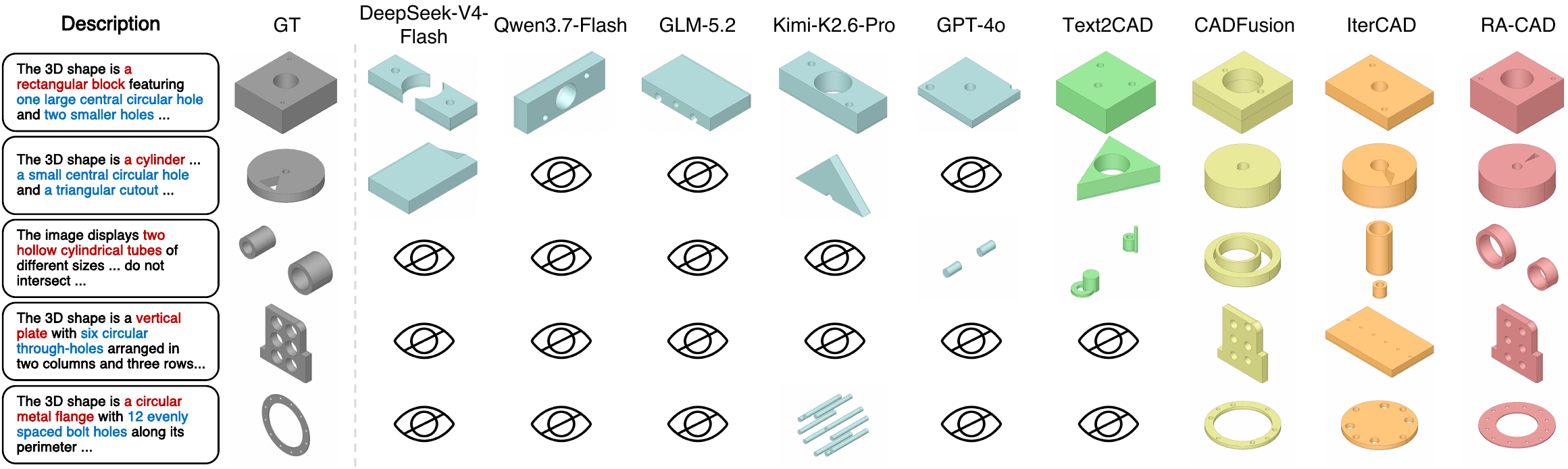} 
\caption{Qualitative comparison of baseline methods and different model variants under different training policies. Non-executable outputs are marked. The far-right column displays the overall model of our method, which has the highest executability and best preserves the geometry.}
\label{CaseS}
\end{figure*}

\begin{table}[t]
\caption{Comparison of RA-CAD with proprietary LLMs under the same prompting protocol. The results are measured by sequence-level accuracy (Avg F1), geometric similarity (Avg CD), and execution validity (IR).}
\small
\setlength{\tabcolsep}{1mm}
\centering
\begin{tabular}{l|ccc|ccc}
  \toprule
  & \multicolumn{3}{c|}{CADFusion} & \multicolumn{3}{c}{Text2CAD} \\
  \cmidrule(lr){2-4} \cmidrule(lr){5-7}
  Method & Avg F1$\uparrow$ & Avg CD$\downarrow$ & IR$\downarrow$
         & Avg F1$\uparrow$ & Avg CD$\downarrow$ & IR$\downarrow$ \\
  \midrule
  DeepSeek-V4-Flash   & 80.27 & 51.59 & 66.77 & 77.41 & 55.82 & 36.22\\
  Qwen3.7-Flash       & 79.71 & 56.32 & 66.67 & 77.17 & 51.97 & 44.30\\
  GLM-5.2             & 83.36 & 55.48 & 87.80 & 78.73 & 49.56 & 38.31\\
  Kimi-K2.6-Pro       & 80.36 & 41.47 & 66.25 & \textbf{78.78} & 47.69 & 70.27\\
  GPT-4o-mini         & 81.92 & 66.73 & 69.72 & 77.94 & 56.53 & 70.25\\
  GPT-4o              & 81.86 & 61.53 & 54.26 & 78.12 & 50.16 & 63.29\\
  \midrule
  RA-CAD & \textbf{85.04} & \textbf{29.04} & \textbf{1.05} & 76.57 & \textbf{33.35} & \textbf{0.89} \\
  \bottomrule
\end{tabular}
\label{CloseS}
\end{table}

We further evaluate its performance against several state-of-the-art proprietary LLMs. Specifically, considering that reasoning ability may have an additional impact on the generated results, to ensure comparative fairness, we adopt a uniform 8-shot prompting strategy.
The quantitative results are presented in \Cref{CloseS} and the qualitative results are also shown in \Cref{CaseS}. Although these proprietary models exhibit remarkable capabilities in general-purpose code generation and reasoning, they perform considerably worse on parametric CAD generation. In particular, they frequently generate syntactically invalid CAD codes or geometrically inconsistent modeling operations, resulting in a high IR.

\begin{table*}[t]
\caption{Ablation studies on the effectiveness of different components in RA-CAD on the CADFusion dataset. The experiments evaluate the contributions of CCB and FAO, among which FAO is composed of the Generation/Rewriting and Critique of the agent and GRPO with policy learning for the corresponding modules.}
\small
\centering
\begin{tabular}{l|ccc|cc|ccc}
  \toprule
    Variant & CCB & G\&R & Crit & $\pi_\theta^{\mathrm{g\&r}}$ & $\pi_\theta^{\mathrm{crit}}$ & Avg F1$\uparrow$ & Avg CD$\downarrow$ & IR$\downarrow$ \\
  \midrule
    \textit{w/o} CCB & \ding{55} & \ding{51} & \ding{51} & \ding{51} & \ding{51} & - & - & 100.00 \\
    \textit{w/o} FAO & \ding{51} & \ding{55} & \ding{55} & \ding{55} & \ding{55} & 75.87 & 59.55 & 68.98 \\
    \textit{w/o} Crit & \ding{51} & \ding{51} & \ding{55} & \ding{51} & \ding{55} & 84.93 & 44.24 & 11.46 \\
  \midrule
    \textit{w/o} GRPO & \ding{51} & \ding{51} & \ding{51} & \ding{55} & \ding{55} & 81.92 & 38.00 & 18.82 \\
    \textit{fixed} G\&R & \ding{51} & \ding{51} & \ding{51} & \ding{55} & \ding{51} & 82.19 & 36.97 & 4.42 \\
    \textit{fixed} Crit & \ding{51} & \ding{51} & \ding{51} & \ding{51} & \ding{55} & 83.50 & 32.72 & 7.05 \\
  \midrule
    Full & \ding{51} & \ding{51} & \ding{51} & \ding{51} & \ding{51} & \textbf{85.04} & \textbf{29.04} & \textbf{1.05} \\
  \bottomrule
\end{tabular}
\label{AblationS}
\end{table*}

\subsection{Ablation Studies}
We also isolate the impact of each component in the experiment pipeline, and the final results are shown in \Cref{AblationS}. We first consider that without CCB, the performance of the model would be so poor that even an effective model could not be generated, indicating the necessity of using prior knowledge in reinforcement learning based on execution results. 
Then, we remove the FAO stage and only retain the model trained in the CCB stage. Under this simplified CCB-only configuration, the model's performance indicators indicate that the FAO stage is necessary for the capability to fully capture the spatial reasoning of complex CAD structures.
Similarly, we remove the Critique from the agent and simplify the framework to traditional reinforcement learning without critique-correction generation. Since the Critique module determines whether the generated code should undergo further rewriting, ablating it directly collapses the multi-round interaction into a single-round generation process. 
All evaluation metrics deteriorate compared with the complete RA-CAD. These results suggest that performance gains are not only brought about by CCB and FAO themselves but also largely benefit from the post-execution critique mechanism from the agent.

To further disentangle the respective contributions of the Generation/Rewriting (G\&R) and Critique modules within the agent, we conduct two additional controlled variants by fixing the policy update of one module while keeping the other trainable. Compared with the \textit{w/o} GRPO variant, which removes reinforcement learning entirely, both \textit{fixed} G\&R and \textit{fixed} Critique still achieve notably better performance. Notably, fixing the Critique module results in a better Avg CD than fixing the G\&R module, while fixing the G\&R module yields a better IR, implying that the two modules play asymmetric roles: the G\&R training is more directly responsible for geometric fidelity, while Critique training is more directly responsible for execution validity.
These results confirm that jointly optimizing both modules is essential, as the co-evolution of code generation and critique provides complementary learning signals that neither module can obtain when optimized in isolation.

Overall, the complete RA-CAD achieved the best performance in all evaluation metrics. These results indicate that these components complement each other and contribute to the outstanding performance of RA-CAD.

\section{Conclusion}
In this paper, we presented a ReAct Agent for CAD generation (RA-CAD), which learned not only to generate CAD codes but also to critique and revise them after execution. Its central formulation factorized a complete code proposal/rewrite and an explicit post-execution critique within one state-aware policy, then jointly optimized their token sequences over complete trajectories. CCB supplied executable CAD priors, and FAO used terminal sequence and geometry quality to improve critique-guided rewriting without a separate value critic or reward model. The reported results and ablations indicated improved geometric fidelity and validity ratio on CADFusion and Text2CAD, while leaving direct visual conditioning, per-round convergence, and broader code-space generalization for future evaluation.

\bibliography{iclr2027_conference}
\bibliographystyle{iclr2027_conference}

\appendix

\newpage
\section{Description}

\begin{listing}[t]
\begin{lstlisting}[]
@\mbox{\textbf{"Description":}}@ "@\mbox{\textbf{[Shape Overview]}}@ The 3D shape consists of a rectangular prism extending into a semi-cylinder with a curved end. @\mbox{\textbf{[Shape Details]}}@ It features one large circular hole on the rectangular part and four smaller, equally sized holes aligned along the semi-cylindrical section. @\mbox{\textbf{[Shape Applications]}}@ The shape is symmetrically aligned along its central axis and potentially designed for mounting or connecting with other components."
\end{lstlisting}
\caption{A representative sample of CADFusion's description structures. The three-stage structure has been highlighted in bold font.}
\label{cadfusion}
\end{listing}

\begin{listing}[t]
\begin{lstlisting}[]
{
  @\mbox{\textbf{"L0":}}@ "A rectangular prism with a cylindrical hole in the center. The prism has four circular cutouts on its top surface."
  @\mbox{\textbf{"L1":}}@ "The CAD model features a rectangular bar with a cylindrical hole in the center and four circular cutout sections on top."
  @\mbox{\textbf{"L2":}}@ "The CAD model consists of a rectangular prism with a cylindrical hole in the center and circular cutouts on the top surface. This part is created by sketching inside a coordinate system and extruding and cutting shapes in 3D. The resulting part has a height of approximately 0.2 units."
  @\mbox{\textbf{"L3":}}@ "The first part of the CAD model is a rectangular prism with a cylindrical hole in the center and four circular cutouts on the top surface. Begin by creating a new coordinate system with Euler angles of [0.0, 0.0, 0.0] and a translation vector of [0.0, 0.0, 0.0]. In the first 2D sketch, draw a rectangle by creating four lines with the following endpoints:* Line 1: [0.0, 0.0] and [0.3, 0.0] * Line 2: [0.3, 0.0] and [0.3, 0.3] ......"
}
\end{lstlisting}
\caption{A representative sample of CADFusion's description structures. It can be seen that the descriptions from L0 to L3 become increasingly detailed. The description of L3 is quite detailed, involving multiple curves and instructions for Boolean operations. Due to the excessive amount of content, it has been omitted for the time being.}
\label{text2cad}
\end{listing}

In the text-to-CAD generation task, the format and granularity of the input text description have a significant impact on the learning efficiency and generation quality of the model. The description structures of the two datasets we selected follow the following rules.

CADFusion~\citep{wang2025texttocad} adopts a three-stage structured description paradigm, dividing the text expression of the CAD model into three levels: \textbf{(1) Shape Overview}, which is used to describe the overall geometric configuration and macroscopic features; \textbf{(2) Shape Details}, covering specific geometric attributes such as size ratios, component quantities, and spatial arrangement; \textbf{(3) Shape Applications}, which clarify the potential functions or usage scenarios of this shape. As shown in \Cref{cadfusion}, Shape Overview is a mandatory part, while the details and applications parts can be selected flexibly according to the complexity of the model.

Text2CAD~\citep{khan2024text2cad} establishes a four-level progressive description system, as shown in \Cref{text2cad}, dividing the information granularity and professional level from simple to complex into four grades:
\begin{itemize}
\item \textbf{L0 (Abstract Level):} It automatically describes the overall shape of the CAD model from a visual perspective, focusing on category generalization.
\item \textbf{L1 (Beginner Level):} It simplifies the description of design steps using non-professional terms, avoiding complex parameters and measurement information, suitable for preliminary design or non-professional users.
\item \textbf{L2 (Intermediate Level):} While maintaining readability, introduces generalized descriptions of geometric shapes, balancing technical accuracy and expression universality.
\item \textbf{L3 (Expert Level):} Provides precise geometric parameters and relative measurement information, including numerical descriptions such as coordinates, dimensions, and directions, meeting the refined modeling needs of professional designers.
\end{itemize}
Considering that user descriptions in actual CAD design scenarios typically focus on high-level semantics such as part functionality, structural features, and overall geometric attributes, it is difficult to accurately specify the specific geometric parameters corresponding to each edge and arc. Moreover, due to the different construction of CAD code sequences, when converting the Text2CAD dataset, we mainly consider the L2 level description.

\begin{figure*}[t]
\centering
\includegraphics[width=\textwidth]{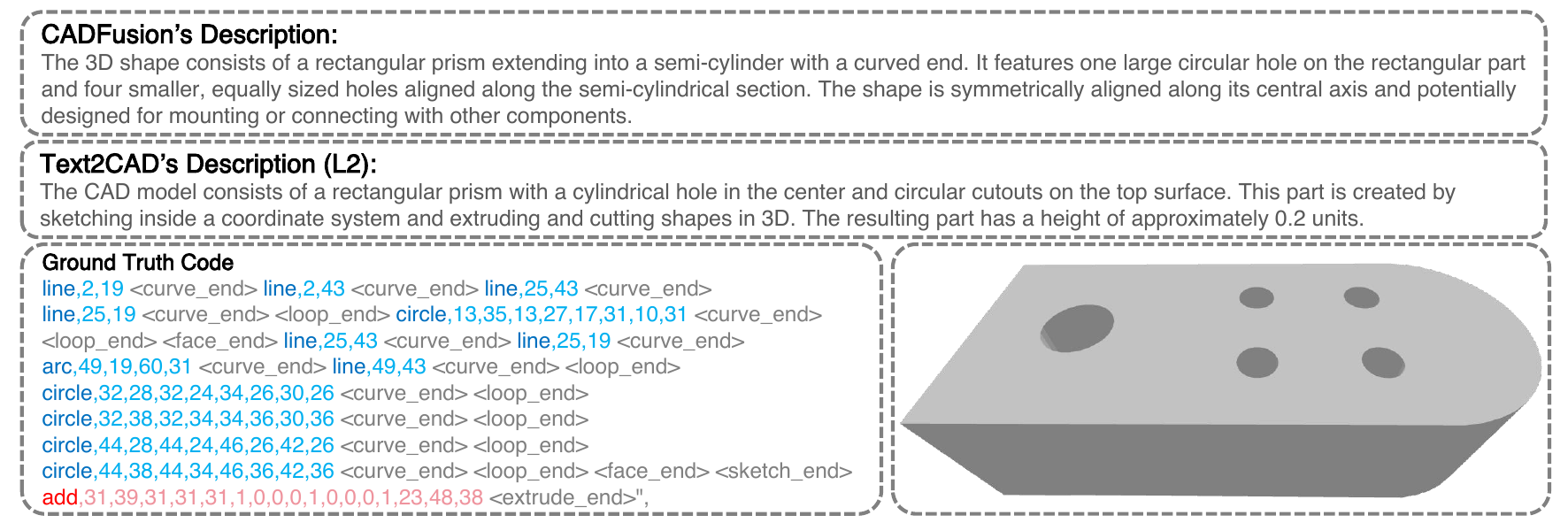} 
\caption{A representative sample of CAD code and its corresponding descriptions. The upper and middle areas respectively represent the descriptions of CADFusion and Text2CAD. The lower left area represents the CAD code sequence, where all types of tokens are distinguished by different colors. All of them correspond to the same CAD model, which is displayed in the lower right corner.}
\label{CodeC}
\end{figure*}

\section{CAD Code}
To achieve the structured modeling and editability of CAD models, this paper adopts a serialized representation format based on "Sketch-and-Extrude". This sequence rule is derived from SkexGen~\citep{xu2022skexgen} and is further standardized by CADFusion into a unified textual CAD language. Each CAD instance is composed of several sketches (Sketch) and extrusion operations in the order of design history. The specific coding rules are as follows:
\begin{itemize}
\item \textbf{Topological Tokens:} Topological Tokens are used to identify the basic geometric elements in 2D sketches, including three categories: $\{\texttt{line}, \texttt{arc}, \texttt{circle}\}$, representing the line primitive, arc primitive, and circle primitive, respectively.
\item \textbf{Topological Parameter Tokens:} Each type of Topological Token must be followed by a different number of coordinate parameters, in the following format: (1) $\texttt{line}, x_0, y_0$; (2) $\texttt{arc}, x_0, y_0, x_1, y_1$; (3) $\texttt{circle}, x_0, y_0, x_1, y_1, x_2, y_2, x_3, y_3$. Among these, $x$ and $y$ represent the horizontal and vertical coordinates, respectively, and $(x_0, y_0)$ represents the starting point coordinates of the curve.
\item \textbf{Extrusion Tokens:} Extrusion Tokens are used to define the types of volume operations in 3D modeling, including three types of Boolean operations: $\{\texttt{add}, \texttt{cut}, \texttt{intersect}\}$, representing adding the new entity to the existing entity (which can be empty), cutting the corresponding entity from the existing entity, and intersecting the new entity with the existing entity, respectively.
\item \textbf{Extrusion Parameter Tokens:} Each type of Extrusion Token must be followed by a total of 17 numerical parameters, which are organized in order as follows: (1) Parameters 1–2: $ext\_v$ (extrusion height on the upper and lower planes); (2) Parameters 3-5: $ext\_T$ (3D extrusion translation vector); (3) Parameters 6-14: $ext\_R$ (3*3 rotation matrix); (4) Parameter 15: $scale\_quan$ (quantized value of the normalized scale); (5) Parameters 16-17: $offset\_quan$ (quantized value of the center offset).
\item \textbf{Hierarchical End Tokens:} To clearly define the multi-level nested boundaries of the CAD structure, the following Hierarchical End Tokens are defined: $\{$$\texttt{<curve\_end>},$ $\texttt{<loop\_end>},$ $\texttt{<face\_end>},$ $\texttt{<sketch\_end>},$ $\texttt{<extrude\_end>}$$\}$, marking the end of the current curve, loop, face, sketch, and extrusion, respectively.
\end{itemize}

Among them, each Topological Parameter Token only defines the starting point of the Topological Token and the intermediate control points along the path, while the coordinate of the endpoint is implicitly determined by the starting point of the next curve. When it is necessary to close the current loop, the endpoint of the last curve points back to the starting point of the first curve in the loop, thus achieving the closure of the loop. For sketches, each sketch is composed of multiple faces, and each face contains one or more loops. The first loop defines the external boundary, and subsequent loops define the internal holes. This hierarchical constraint ensures that the generated CAD sequence is geometrically and topologically legally closed. All coordinate parameters and continuous parameters are uniformly quantized into 6-bit discrete Tokens (with a total of 64 possible values), enabling the entire CAD sequence to be input into the LLM in a unified text Token form for processing.

\section{Prompt}
\subsection{CCB Prompt}

\begin{listing}[t]
\begin{lstlisting}[]
Below is a description of a 3D shape:  
@\color{red}\{Description\}@
Generate a Computer-Aided Design(CAD) command sequence of the 3D shape:
@\color{blue}\{Ground\_Truth\_Code[:i]\}@
\end{lstlisting}
\caption{Prompt of the SFT training format. The model receives the full description and partial code sequence as input, while the loss is computed only on the next tokens (the part after the prompt).}
\label{sft}
\end{listing}

As the first training stage of RA-CAD, CCB initializes the agent's parametric CAD coding ability through supervised fine-tuning. The SFT prompt format is shown in \Cref{sft}. Each training sample is an instruction-style prompt that first presents the natural-language description of the target shape and then requests the corresponding CAD command sequence. To train the model in a next-token prediction manner, the prompt is constructed with only the first \(i\) tokens of the ground-truth code ({Ground\_Truth\_Code[:i]}), and the cross-entropy loss is computed exclusively on the subsequent tokens that follow the prompt. This formulation teaches the model to continue a partially generated CAD sequence rather than to re-predict the prefix it has already observed. At the same time, all training samples share the same description format and the same token-level CAD code used in the subsequent FAO stage.

\subsection{FAO Prompt}

\begin{listing}[t]
\begin{lstlisting}[]
@\mbox{\textbf{"system":}}@"
  You are a professional CAD code generation expert, using token sequences to represent CAD models.
  Based on the user's description of the CAD model, directly generate a valid token sequence. Strictly adhere to the continuous contour and hierarchical structure.
  ",
@\mbox{\textbf{"user":}}@"
  Below is a description of a 3D shape:
  @\color{red}\{Description\}@
  Generate a Computer-Aided Design(CAD) command sequence of the 3D shape:
  "
\end{lstlisting}
\caption{Prompt of the Generation module.}
\label{gen}
\end{listing}

\begin{listing}[t]
\begin{lstlisting}[]
@\mbox{\textbf{"system":}}@"
  You are a professional CAD code regeneration expert, using token sequences to represent CAD models.
  Rewrite the CAD code sequence based on the feedback.
  Please fix the issues in the code and generate a correct compact format sequence.
  ",
@\mbox{\textbf{"user":}}@"
  Below is a description of a 3D shape:
  @\color{red}\{Description\}@
  Original code:
  @\color{blue}\{Code\}@
  Execution result:
  @\color{purple}\{Execution\}@
  Feedback: 
  @\color{green}\{Feedback\}@
  Please regenerate a Computer-Aided Design(CAD) command sequence of the 3D shape:
  "
\end{lstlisting}
\caption{Prompt of the Rewriting module.}
\label{rew}
\end{listing}

\begin{listing}[t]
\begin{lstlisting}[]
@\mbox{\textbf{Input:}}@
@\color{blue}\{CAD\_Code\}@

@\mbox{\textbf{Output - Success:}}@
  Code executed successfully. The generated model is structurally sound.
  Validation passed - {L} token(s), {I} sketch/extrude pair(s),
  {J} face(s), {K} loop(s), {H} curve(s)

@\mbox{\textbf{Output - Failure:}}@
  Code execution failed with error: Cannot be parsed or rendered into an actual 3D model.
  Found {N} error(s):
  1.Sketch-Extrude {i} Face {j} Loop {k} Curve {h}: {specific error}
  2. ...
\end{lstlisting}
\caption{Execution module. The execution module returns either a success message with structural validation statistics, or a failure message containing the exact error location and description.}
\label{exe}
\end{listing}

\begin{listing*}[t]
\begin{lstlisting}[]
@\mbox{\textbf{"system":}}@"
  You are a CAD code critique expert who pays great attention to details of the following tokens that represent CAD models.
  @\mbox{\textbf{Token types:}}@
  1) Topological tokens: line, arc, circle
  2) Extrusion tokens: add, cut, intersect
  3) Hierarchical end tokens: <curve_end>, <loop_end>, <face_end>, <sketch_end>, @\mbox{<extrude\_end>}@
  Please review the current CAD code item by item from the following three dimensions:
  @\mbox{\textbf{1. Program Executability}}@
  -Can the code be compiled and run without errors through the CAD execution environment?
  -If the execution fails, what is the specific error type and on which line/parameter did the error occur?
  @\mbox{\textbf{2. Design Intent Consistency}}@
  -Does the generated CAD model fully cover all geometric features and functional requirements in the design description?
  -Are the features (stretch, rotation, hole, chamfer, array, etc.) appearing in the correct position and size?
  -Are there any missing or redundant features that do not conform to the design description?
  @\mbox{\textbf{3. Geometric Reasonability}}@
  -Is the overall scale of the model coordinated? Is there a clearly unreasonable aspect ratio or wall thickness?
  -Are there any unexpected intersections, penetrations, or gaps between geometric entities?
  -Is there degenerate geometry (zero volume, coplanarity, self intersection, etc.)?
  ### @\mbox{\textbf{Output Format}}@ ###
  You always first perform your validation reasoning by wrapping your analysis in <think> and </think>. In your reasoning, examine the code item by item against the specification and the execution result.
  After completing your reasoning, provide your final validation result within <result> and </result> tags:
  - If errors are found: A bulleted list of specific issues, each on a new line.
  - If no errors are found: The exact phrase "@\mbox{\textbf{THE CODE IS CORRECT}}@".
  ",
@\mbox{\textbf{"user":}}@"
  CAD description: 
  @\color{red}\{Description\}@
  Generated code:
  @\color{blue}\{Code\}@
  Execution result:
  @\color{purple}\{Execution\}@
  "
\end{lstlisting}
\caption{Prompt of the Critique module.}
\label{crit}
\end{listing*}

During the FAO stage, the agent is organized as a Generate–Execute–Critique–Rewrite loop, in which each module is driven by a dedicated role-specific prompt. 
The Generation module, as shown in \Cref{gen}, is instructed as a professional CAD code generation expert and is required to directly produce a valid token sequence while strictly adhering to the continuous contour and hierarchical structure of the CAD language. 
The Rewriting module, as shown in \Cref{rew}, is prompted as a professional CAD code regeneration expert and takes the description, the original code, the execution result, and the critique feedback as input, with the instruction to fix the identified issues and regenerate a correct and compact sequence. 
The Execution module acts as the bridge between the agent and the CAD environment. As shown in \Cref{exe}, it receives the CAD code generated or rewritten by the Generation/Rewriting modules, parses the token sequence, validates the hierarchical structure, and instantiates the corresponding 3D model. When execution succeeds, it returns a success message together with the structural validation statistics, including the total number of tokens, sketch/extrude pairs, faces, loops, and curves. When execution fails, it returns a failure message and reports the specific error location, such as the affected Sketch-Extrude, Face, Loop, and Curve indices, so that the agent can identify exactly where the problem occurred.
The Critique module receives the CAD description, the generated code, and its execution result. As shown in \Cref{crit}, its system prompt enumerates the topological tokens, extrusion tokens, and hierarchical end tokens of the CAD representation, and asks the model to review the code item by item along three dimensions: program executability, design intent consistency, and geometric reasonability. Its output format requires the model to first perform validation reasoning and then return a structured verdict, which standardizes the feedback consumed by the subsequent module.
The same prompts are reused consistently throughout FAO training and inference. During GRPO, each complete interaction trajectory is produced by rolling out these four modules in turn against the real CAD execution environment, and all policy-generated code and critique tokens are jointly optimized under terminal F1 and Chamfer Distance rewards. The role prompts thus provide a stable interface between the environment, the critique decision, and the rewrite, ensuring that the policy learns to turn execution feedback into explicit corrective actions rather than relying on fixed or unoptimized textual guidance.

\subsection{Few-shot Prompt}

\begin{listing*}[t]
\begin{lstlisting}[]
Below is a description of a 3D shape:
@\color{red}\{Description\}@
Generate a Computer-Aided Design (CAD) command sequence of the 3D shape. The command sequence involves sketches such as lines, arcs, and circles, each marked by the endpoints, and extrusions that make the sketch into 3D volumes.
Here are some examples and their value range is 0-63
1. Description: @\color{red}\{Description[0]\}@, CAD Command Sequence: @\color{blue}\{Ground\_Truth\_Code[0]\}@
2. Description: @\color{red}\{Description[1]\}@, CAD Command Sequence: @\color{blue}\{Ground\_Truth\_Code[1]\}@
3. Description: @\color{red}\{Description[2]\}@, CAD Command Sequence: @\color{blue}\{Ground\_Truth\_Code[2]\}@
4. Description: @\color{red}\{Description[3]\}@, CAD Command Sequence: @\color{blue}\{Ground\_Truth\_Code[3]\}@
5. Description: @\color{red}\{Description[4]\}@, CAD Command Sequence: @\color{blue}\{Ground\_Truth\_Code[4]\}@
6. Description: @\color{red}\{Description[5]\}@, CAD Command Sequence: @\color{blue}\{Ground\_Truth\_Code[5]\}@
7. Description: @\color{red}\{Description[6]\}@, CAD Command Sequence: @\color{blue}\{Ground\_Truth\_Code[6]\}@
8. Description: @\color{red}\{Description[7]\}@, CAD Command Sequence: @\color{blue}\{Ground\_Truth\_Code[7]\}@
Now it's your turn. Remind that this is your description: @\color{red}\{Description\}@.
No explanation is needed. Only return your final sequence, and in one line.
\end{lstlisting}
\caption{Prompt of the few-shot. For different datasets, we respectively selected 8 distinct samples as the few-shot examples.}
\label{fews}
\end{listing*}

To evaluate RA-CAD against strong proprietary language models under a fair and reproducible protocol, we use the few-shot prompt shown in \Cref{fews} for all closed-source baselines. The prompt first states the task of generating a CAD command sequence from a shape description and briefly explains the structure of the sequence, namely sketches composed of lines, arcs, and circles together with the extrusions that turn a sketch into a 3D volume. It then provides eight complete description-code demonstration pairs and explicitly indicates that the tokenized values range from 0 to 63, which corresponds to the 6-bit quantization of all coordinate and continuous parameters. Finally, the model is asked to generate the sequence for the target description without providing any explanation and to return only the final sequence in a single line.

Considering that there are differences in the description methods between the CADFusion and Text2CAD datasets, we selected 8 representative samples from each of the two datasets as the few-shot examples, as shown in \Cref{fuss} and \Cref{t2cs} respectively. When selecting, we took into account the difficulty gradient and structural complexity of the samples, and ensured that the examples covered as many diverse types of CAD tokens as possible, thereby more comprehensively reflecting the generation characteristics of the corresponding datasets.

\begin{table*}[t]
\centering
\small
\renewcommand{\arraystretch}{1.3}
\setlength{\tabcolsep}{7pt}
\begin{tabularx}{\textwidth}{@{} l >{\raggedright\arraybackslash}X c c @{}}
\toprule
\textbf{ID} & \textbf{CAD Model Structure} & \textbf{Topological Tokens} & \textbf{Extrusion Tokens} \\
\midrule
\texttt{0002/00027926} & Rectangular prism
  & \shortstack{\textbf{4}\\line $\times$ 4}
  & \shortstack{\textbf{1}\\add $\times$ 1} \\
\texttt{0034/00343336} & Triangular prism
  & \shortstack{\textbf{3}\\line $\times$ 3}
  & \shortstack{\textbf{1}\\add $\times$ 1} \\
\texttt{0030/00302523} & Rectangular prism + 2 cylindrical holes
  & \shortstack{\textbf{6}\\line $\times$ 4, circle $\times$ 2}
  & \shortstack{\textbf{1}\\add $\times$ 1} \\
\texttt{0002/00023779} & Disc + Center boss + 5 cylindrical holes
  & \shortstack{\textbf{23}\\circle $\times$ 13, arc $\times$ 10}
  & \shortstack{\textbf{2}\\add $\times$ 2} \\
\texttt{0021/00218782} & Rectangular prism + Semi-cylindrical top + Cylindrical hole
  & \shortstack{\textbf{6}\\line $\times$ 4, arc $\times$ 1, circle $\times$ 1}
  & \shortstack{\textbf{1}\\add $\times$ 1} \\
\texttt{0034/00344524} & Rectangular prism + Cutout square hole
  & \shortstack{\textbf{8}\\line $\times$ 8}
  & \shortstack{\textbf{2}\\add $\times$ 1, cut $\times$ 1} \\
\texttt{0027/00272814} & L-shaped prism + Inner corner rounded
  & \shortstack{\textbf{5}\\line $\times$ 4, arc $\times$ 1}
  & \shortstack{\textbf{1}\\add $\times$ 1} \\
\texttt{0025/00250827} & Rectangular prism + Center square hole + Center boss + Center circular hole
  & \shortstack{\textbf{13}\\line $\times$ 12, circle $\times$ 1}
  & \shortstack{\textbf{2}\\add $\times$ 2} \\
\bottomrule
\end{tabularx}
\caption{Few-shot samples from the CADFusion dataset. We used the descriptions and true values of these eight samples as the input for the closed-source model for its reasoning.}
\label{fuss}
\end{table*}

\begin{table*}
\centering
\small
\renewcommand{\arraystretch}{1.3}
\setlength{\tabcolsep}{7pt}
\begin{tabularx}{\textwidth}{@{} l >{\raggedright\arraybackslash}X c c @{}}
\toprule
\textbf{ID} & \textbf{CAD Model Structure} & \textbf{Topological Tokens} & \textbf{Extrusion Tokens} \\
\midrule
\texttt{0000/00000894} & Rectangular prism
  & \shortstack{\textbf{4}\\line $\times$ 4}
  & \shortstack{\textbf{1}\\add $\times$ 1} \\
\texttt{0000/00001260} & Cylinder
  & \shortstack{\textbf{1}\\circle $\times$ 1}
  & \shortstack{\textbf{1}\\add $\times$ 1} \\
\texttt{0001/00016772} & Rectangular prism + Rounded corners + Curved contour
  & \shortstack{\textbf{9}\\line $\times$ 5, arc $\times$ 4}
  & \shortstack{\textbf{1}\\add $\times$ 1} \\
\texttt{0001/00019015} & Rectangular prism + Cylinder
  & \shortstack{\textbf{8}\\line $\times$ 8}
  & \shortstack{\textbf{2}\\add $\times$ 2} \\
\texttt{0000/00002694} & Rectangular prism + Central circular hole + Cutout circular pit + Rectangular prism
  & \shortstack{\textbf{13}\\line $\times$ 8, circle $\times$ 5}
  & \shortstack{\textbf{3}\\add $\times$ 2, cut $\times$ 1} \\
\texttt{0055/00552776} & Rectangular prism + Central square hole + Cylinder
  & \shortstack{\textbf{12}\\line $\times$ 12}
  & \shortstack{\textbf{2}\\add $\times$ 2} \\
\texttt{0057/00578900} & T-shaped prism + 5 cylindrical holes
  & \shortstack{\textbf{15}\\line $\times$ 10, circle $\times$ 5}
  & \shortstack{\textbf{1}\\add $\times$ 1} \\
\texttt{0055/00558114} & Complex multi-component assembly
  & \shortstack{\textbf{18}\\line $\times$ 10, arc $\times$ 3, circle $\times$ 5}
  & \shortstack{\textbf{4}\\add $\times$ 4} \\
\bottomrule
\end{tabularx}
\caption{Few-shot samples from the Text2CAD dataset. We used the descriptions and true values of these eight samples as the input for the closed-source model for its reasoning.}
\label{t2cs}
\end{table*}

\section{Implementation Details}

The FAO stage is implemented with the verl framework on a single node with two GPUs, and the agent is initialized from Meta-Llama-3-8B-Instruct. Rollouts are generated with the asynchronous vLLM engine at a temperature of 0.7, sampling \(n=8\) candidate trajectories per prompt; prompts and per-turn responses are both capped at 1,024 tokens, respectively, and truncation is disabled so that any over-length sequence raises an error instead of being silently cut. Since the agent operates as a Generate–Execute–Critique–Rewrite loop, multi-turn rollouts use the Hermes tool-calling format, which allows each module's structured output to be parsed and routed through the CAD execution environment. Each training step samples 16 prompts (128 trajectories in total), with a mini-batch size of 8 and a micro-batch size of 4 per GPU. The policy and reference models are trained with FSDP and parameter offloading; gradient checkpointing is enabled, and the reference model is kept frozen for computing reference log probabilities. GRPO is applied with group-normalized advantages, a fixed KL penalty with coefficient 0.001 incorporated into the reward, policy clipping ratios of 0.2 and 0.3, and a learning rate of \(10^{-6}\), for a single epoch over the training prompts.

As for the assessment, generated CAD sequences are first parsed into sketch-and-extrude operations using the same 6-bit quantization as training, where each quantized value \(q\) is dequantized as 
\[
\hat{x} = q \cdot (\max-\min)/(2^6-1) + \min,
\] 
with sketch coordinates in \([-1,1]\), extrusion heights and translations in \([-1,1]\), scale in \([0,1.4]\), and center offset in \([-0.9,0.9]\). A sequence is counted as valid only if it parses into a consistent sequence of sketches and extrusions, passes the structural checks of the parser (correct token counts, closed loops, non-degenerate curves, and valid Boolean operations), and can be rendered through the CAD execution environment. The invalidity ratio is computed as 
\[
IR = (1 - N_{\text{valid}} / N_{\text{total}}) \times 100\%.
\] 
For geometric evaluation, 2,000 points are sampled from the surface of each rendered mesh (downsampled from denser point clouds when necessary) and normalized by the maximum absolute coordinate. The Chamfer distance between a generated point cloud \(P\) and its ground-truth point cloud \(Q\) is defined as 
\[
CD(P, Q) = \frac{1}{|P|}\sum_{p \in P}\min_{q \in Q}\|p-q\|_2^2 + \frac{1}{|Q|}\sum_{q \in Q}\min_{p \in P}\|p-q\|_2^2
\]
and the reported Avg CD is the mean of the diagonal entries of the pairwise distance matrix, i.e., each generated model is compared only with its corresponding ground-truth model. For sequence-level accuracy, F1 scores are computed per primitive type (line, arc, and circle) within each sketch-extrude segment: for type \(t\), with occurrence counts \(c_t^{\text{gt}}\) and \(c_t^{\text{pred}}\),
\[m_t = \min(c_t^{\text{gt}}, c_t^{\text{pred}}), \]
\[\text{Precision}_t = m_t / c_t^{\text{pred}},\] 
\[\text{Recall}_t = m_t / c_t^{\text{gt}},\] 
\[F1_t = 2 \cdot \text{Precision}_t \cdot \text{Recall}_t / (\text{Precision}_t + \text{Recall}_t), \]
where per-type scores are averaged over the evaluated samples, and Avg F1 is the mean of the line, arc, and circle F1 scores. The same pipeline additionally computes the Jensen–Shannon divergence (JSD) between the per-cell occupancy probabilities of the generated and ground-truth point clouds on a \(28 \times 28 \times 28\) grid in the unit cube.

\section{Case Studies}

\begin{figure*}[t]
\centering
\includegraphics[width=0.8\textwidth]{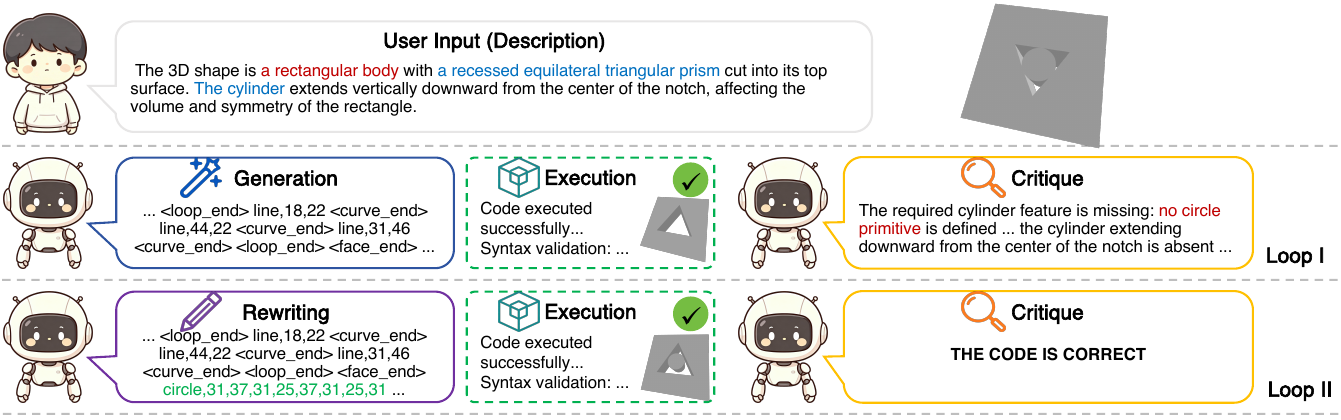} 
\caption{The text-to-CAD generation process case. The first loop model is missing a cylinder, while the second loop adds the circle feature for correction.}
\label{t1}
\end{figure*}

\begin{figure*}[t]
\centering
\includegraphics[width=0.8\textwidth]{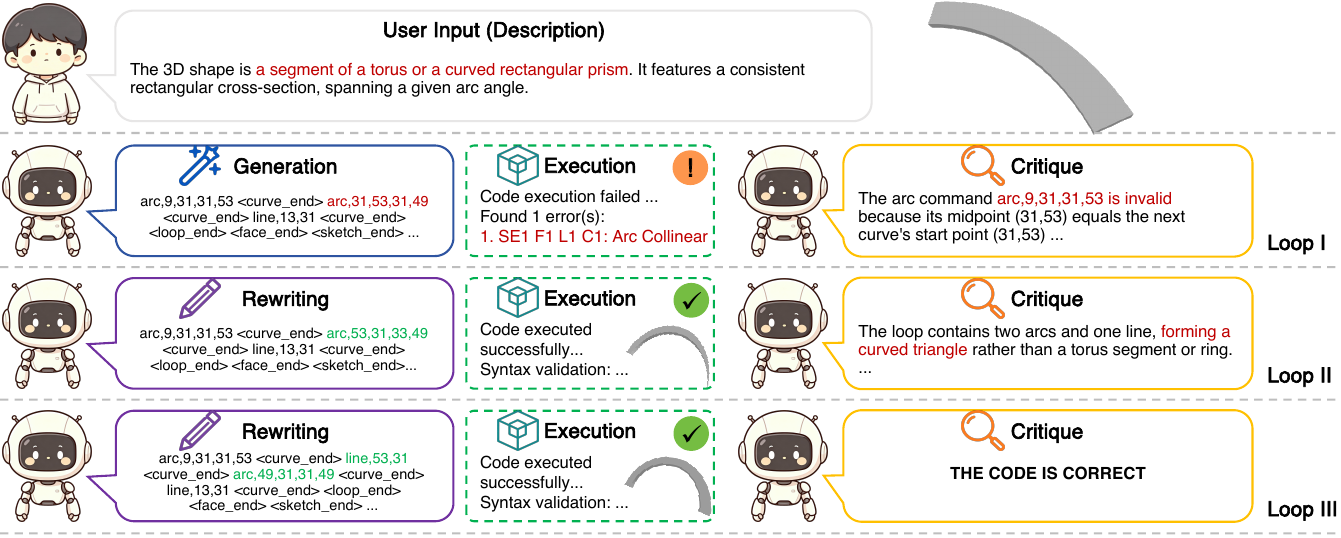} 
\caption{The text-to-CAD generation process case. The first loop fails execution due to an arc with duplicate points, the second loop executes but forms a curved triangle rather than a ring, and the third loop correctly builds the shape from two concentric arcs and two straight sides.}
\label{t2}
\end{figure*}

We further provide qualitative case studies to illustrate how RA-CAD improves CAD code in the closed-loop Generate–Execute–Critique–Rewrite paradigm using execution feedback and explicit critique. \Cref{t1} and \Cref{t2} show the text-to-CAD generation process: the first process shows how RA-CAD corrects a missing structural feature through feedback-driven rewriting, while the second process shows how RA-CAD leverages the detailed failure messages returned by the CAD execution environment to locate and repair errors.

\begin{figure*}[t]
\centering
\includegraphics[width=\textwidth]{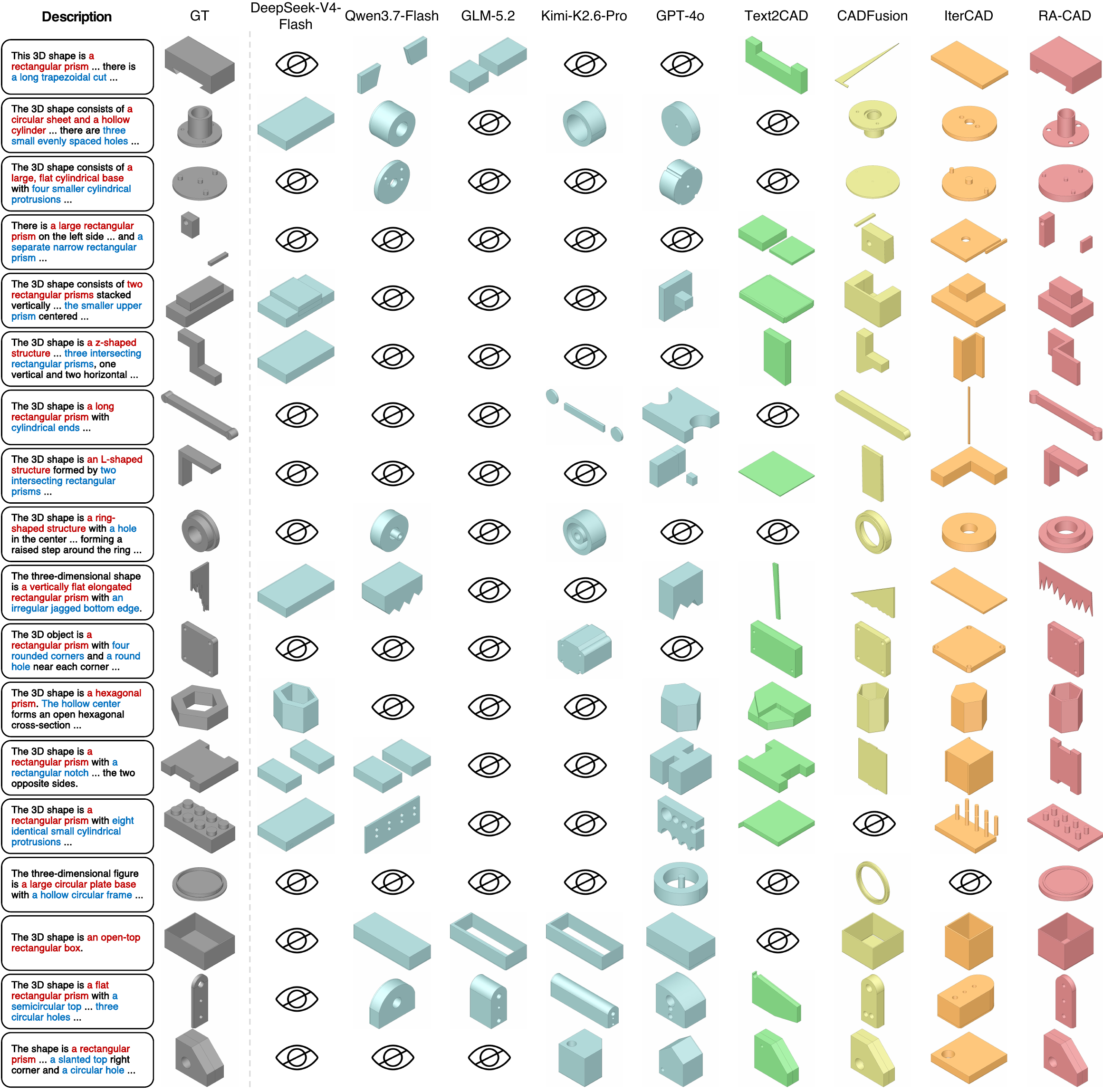} 
\caption{Qualitative comparison of baseline methods and different LLM variants under different training policies. Non-executable outputs are marked.}
\label{Cases}
\end{figure*}

\Cref{Cases} further presents qualitative comparisons of the final outputs on several representative cases. Compared with strong proprietary models (DeepSeek-V4-Flash, Qwen3.7-Flash, GLM-5.2, Kimi-K2.6-Pro, and GPT-4o) and existing methods (Text2CAD and CADFusion), RA-CAD best preserves the requested structures and geometric relations while remaining executable, whereas most baselines produce non-executable or geometrically inconsistent outputs.

\section{Comparison Protocol and Reproducibility Constraints}
Several recent text-to-CAD and agent-based methods, including CAD-coder~\citep{guan2025cadcoder}, CADDesigner~\citep{fan2026caddesigner}, PR-CAD~\citep{an2026prcad}, ReCAD~\citep{li2026recad}, ToolCAD~\citep{gong2026toolcad}, CME-CAD~\citep{niu2026cmecad}, and CAD-Factory~\citep{liu2026cadfactory}, are not publicly reproducible at the time of submission: their source code, trained checkpoints, or complete evaluation pipelines are unavailable. In addition, some of them use different CAD representations (e.g., B-Rep, mesh, or alternative parametric DSLs), input modalities (e.g., images, sketches, or point clouds), datasets, or evaluation protocols, which prevents a faithful apples-to-apples comparison under our unified sketch-and-extrude code representation and terminal CAD-quality metrics.

To ensure that all reported comparisons are based on identical data preprocessing, execution environments, and evaluation metrics, we therefore compare against methods for which implementations and evaluation protocols are publicly accessible, including Text2CAD, CADFusion, and IterCAD. We further include strong proprietary LLMs under a fixed 8-shot prompting protocol as an additional reference point. This comparison protocol does not imply universal superiority over methods that could not be reproduced under the same setting; rather, it isolates the contribution of RA-CAD under a controlled and reproducible evaluation.

We evaluated the included baselines on the same processed test splits and with the same CAD parser, execution environment, and metric implementation used for RA-CAD. For the baselines, we directly used the checkpoint released with version 1.0 for CADFusion and reimplemented Text2CAD and IterCAD following their respective papers, with IterCAD's visual inputs converted into textual descriptions to match our backbone, and we report our re-evaluated results rather than numbers copied from the original papers. We preserved each method’s original model architecture and training objective where applicable. For one-shot baselines, Single uses one generated sequence. Repeat-5 uses up to five independent generations and selects the first executable sequence; if none of the five candidates were executable, we used the CAD code generated in the fifth attempt as the final output. For iterative baselines, Turns-5 allows at most five generation–execution turns, with each method following the stopping and output-selection rules specified by its corresponding paper. These settings standardize the output evaluation, but do not make the methods identical in input information or inference budget. We therefore interpret the comparison within the stated protocol rather than as a reproduction of every method’s original benchmark setting.

For proprietary LLMs, we report single-sample results under a fixed 8-shot protocol. We did not perform repeated sampling for these models because our controlled Repeat-5 evaluations of the reproducible baselines showed that repeated sampling primarily reduced the invalidity ratio, while changes in Avg F1 and Avg CD were comparatively small across the two datasets. We therefore use single-sample results to compare per-call generation quality under a fixed inference budget. This choice does not imply that repeated sampling would have no effect on proprietary models; in particular, their execution-validity rates may improve with additional samples. Therefore, the comparison of proprietary models in this article should be understood as a comparison of the quality of a single call under a fixed single-sample protocol; the reported results of RA-CAD are also based on a single interaction trajectory, without using the Repeat-N selection to ensure protocol consistency.

\section{Limitations and Future Work} 
Despite the promising results, RA-CAD has several limitations. First, it currently takes only a natural-language description as input, and direct visual conditioning on images, sketches, or point clouds is not supported, which limits its applicability to reverse engineering and visually grounded editing tasks. Second, the code space is restricted to the sketch-and-extrude DSL used by the training datasets, i.e., line, arc, and circle primitives with add/cut/intersect operations and 6-bit quantized parameters. Uncommon industrial operations, as well as continuous-parameter precision, are not covered, so generalization to broader CAD code spaces remains an open problem.

In future work, we plan to extend RA-CAD to multimodal inputs and interactive user feedback, enabling the agent to condition on images, sketches, or point clouds and to incorporate user edits during the design session. We will broaden the supported CAD code space to more diverse operations and finer quantization, and adapt the framework to additional execution environments and datasets.

\end{document}